%% file: main.tex
\documentclass{article} 
\usepackage{iclr2021_conference,times}

\input{math_commands.tex}

\usepackage{hyperref}
\usepackage{url}

\usepackage{tikz}
\usetikzlibrary{arrows,automata,shapes,positioning}
\usepackage{float}
\usepackage{wrapfig}
\usepackage{xcolor}
\usepackage{longtable,lscape}
\usepackage{multirow}
\usepackage{microtype}
\usepackage{graphicx}
\usepackage{subfigure}
\usepackage{booktabs} 
\usepackage{multirow}
\usepackage{seqsplit}

\renewcommand{\rv}[2]{\ensuremath{#1 \pm #2}}
\newcommand{\hlv}[1]{\ensuremath{\mathbf{#1}}}

\newif\ifMergeFigs
\MergeFigstrue

\title{Formal Language Constrained \\ Markov Decision Processes}

\author{
  Eleanor Quint\thanks{Correspondence to equint4@huskers.unl.edu}\enspace$^\dagger$, Dong Xu$^\ddagger$, Samuel W. Flint$^\dagger$, Stephen Scott$^\dagger$, Matthew Dwyer$^\ddagger$\\
  University of Nebraska-Lincoln$^\dagger$, University of Virginia$^\ddagger$
}

\iclrfinalcopy 
\begin{document}

\maketitle

\begin{abstract}
In order to satisfy safety conditions, an agent may be constrained from acting freely. A safe controller can be designed a priori if an environment is well understood, but not when learning is employed. In particular, reinforcement learned (RL) controllers require exploration, which can be hazardous in safety critical situations. We study the benefits of giving structure to the constraints of a constrained Markov decision process by specifying them in formal languages as a step towards using safety methods from software engineering and controller synthesis. We instantiate these constraints as finite automata to efficiently recognise constraint violations. Constraint states are then used to augment the underlying MDP state and to learn a dense cost function, easing the problem of quickly learning joint MDP/constraint dynamics. We empirically evaluate the effect of these methods on training a variety of RL algorithms over several constraints specified in Safety Gym, MuJoCo, and Atari environments.
\end{abstract}

\section{Introduction}
\label{introduction}

The ability to impose safety constraints on an agent is
key to the deployment of reinforcement learning (RL) systems in real-world environments~\citep{amodei2016concrete}. Controllers that are derived mathematically typically rely on a full a priori analysis of agent behavior remaining within a pre-defined envelope of safety in order to guarantee safe operation \citep{arechiga2014using}. This approach restricts controllers to pre-defined, analytical operational limits, but allows for verification of safety properties~\citep{DBLP:conf/lics/HuthK97} and satisfaction of software contracts~\citep{helm1990contracts}, which enables their use as a component in larger systems.
By contrast, RL controllers are free to learn control trajectories that better suit their tasks and goals; however, understanding and verifying their safety properties is challenging.
A particular hazard of learning an RL controller is the requirement of exploration in an unknown environment. It is desirable not only to obey constraints in the final policy, but also throughout the exploration and learning process~\citep{ray2019benchmarking}.

The goal of safe operation as an optimization objective is formalized by the constrained Markov decision process (CMDP)~\citep{altman1999constrained}, which adds to a Markov decision process (MDP) a cost signal similar to the reward signal, and poses a constrained optimization problem in which discounted reward is maximized while the total cost must remain below a pre-specified limit per constraint. We use this framework and propose specifying CMDP constraints in formal languages to add useful structure based on expert knowledge, e.g., building sensitivity to proximity into constraints on object collision or converting a non-Markovian constraint into a Markovian one~\citep{detemporal}.

A significant advantage of specifying constraints with formal languages is that they already form a well-developed basis for components of safety-critical systems~\citep{DBLP:conf/lics/HuthK97,DBLP:books/daglib/0007403,kwiatkowska2002prism,baier2003model} and safety properties specified in formal languages can be verified a priori~\citep{kupferman2000automata,bouajjani1997reachability}. Moreover, the recognition problem for many classes of formal languages imposes modest computational requirements, making them suitable for efficient runtime verification~\citep{Chen:2007:MOP}. This allows for low-overhead incorporation of potentially complex constraints into RL training and deployment.

We propose
    (1) a method for posing formal language constraints defined over MDP trajectories as CMDP cost functions;
    (2) augmenting MDP state with constraint automaton state to more explicitly encourage learning of joint MDP/constraint dynamics;
    (3) a method for learning a dense cost function given a sparse cost function from joint MDP/constraint dynamics; 
and (4)
    a method based on constraint structure to dynamically modify the set of available actions to guarantee the prevention of constraint violations.
We validate our methods over a variety of RL algorithms with standard constraints in Safety Gym and hand-built constraints in MuJoCo and Atari environments.

The remainder of this work is organized as follows. Section~\ref{sec:rel_work} presents related work in CMDPs, using expert advice in RL and safety, as well as formal languages in similar settings.
Section~\ref{formal_language_constraint_framework} describes our definition of a formal language-based cost function, as well as how it's employed in state augmentation, cost shaping, and action shaping.
Section~\ref{experiments} details our experimental setup and results and finally, discussion of limitations and future work are located in Section~\ref{sec:discussion}.

\ifMergeFigs
\begin{figure}
\begin{center}
  \begin{tabular}{cc}
    \resizebox{6 cm}{!}{\input{flc-op-through-time}}
    & 
      \input{no-1d-dithering-dfa}
    \\
    (a) & (b)
  \end{tabular}
    \caption{(a) Illustration of the formal language constraint framework operating through time. State is carried forward through time by both the MDP and the recognizer, $D_C$.
(b) No-1D-dithering constraint employed in the Atari and MuJoCo domains: $.^* \, (\ell \, r)^2|(r\,\ell)^2$ (note, all unrepresented transitions return to $q_0$).}
\label{fig:merged}
\end{center}
\vspace*{-0.25in}
\end{figure}
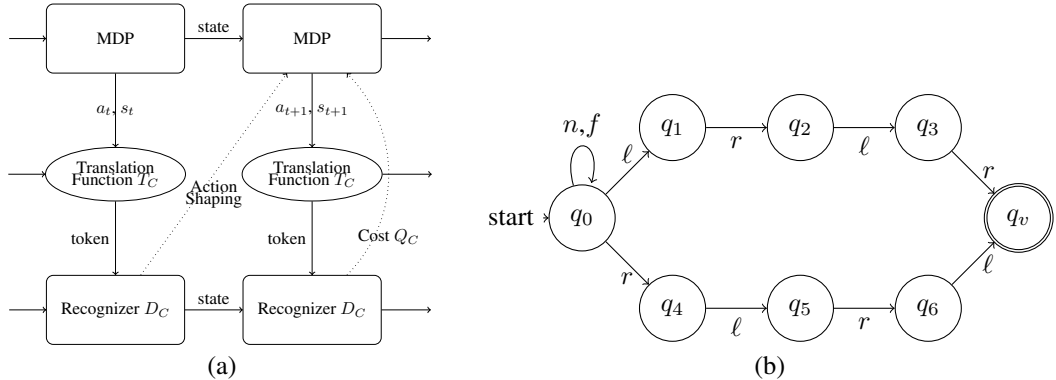
\fi

\unless\ifMergeFigs
\begin{figure}
    \centering
    \includegraphics[width=0.5\textwidth]{constraint_framework.png}
    \vspace{-2mm}
    \caption{Illustration of the formal language constraint framework operating through time. In the same way state is carried forward through time by the MDP, it's also propagated by the translation layer, $\ftl$, and the recognizer, $D$.}
    \label{fig:illustrated_constraint_framework}
\end{figure}
\fi

\unless\ifMergeFigs
\begin{figure}
\begin{center}
\begin{tikzpicture}[->,>=stealth',shorten >=1pt,auto,node distance=2.0cm,
semithick, scale=0.26]
  \tikzstyle{every state}=[draw]

  \node[initial,state] (I)              {$q_0$};
  \node[state]         (L) [right of=I] {$q_1$};
  \node[state]         (LR) [right of=L] {$q_2$};
  \node[state]         (LRL) [right of=LR] {$q_3$};
  \node[state,accepting]         (V) [above of=LRL] {$q_v$};

  \path (I)   edge [loop below] node {$n$,$f$,$r$} (I)
              edge              node [below] {$\ell$} (L)
        (L)   edge [loop below] node {$\ell$} (L)
              edge [bend right=15] node [above] {$n$,$f$} (I)
              edge              node [below] {$r$} (LR)
        (LR)  edge              node [below] {$\ell$} (LRL)
              edge [bend right=45] node [above] {$n$,$f$,$r$} (I)
        (LRL) edge              node {$r$} (V)
              edge [bend right=45] node [above] {$\ell$} (L)
              edge [bend right=60] node [above] {$n$,$f$} (I)
        (V)   edge [loop left]  node {$n$,$f$,$\ell$,$r$} (V);
\end{tikzpicture}
\vspace{-2mm}
\caption{No-1D-dithering constraint: $.^* \, (\ell \, r)^2$\label{fig:nodither}}
\end{center}
\end{figure}
\fi

\section{Related work}
\label{sec:rel_work}

\paragraph{Safety and CMDP Framework} The CMDP framework doesn't prescribe the exact form of constraints or how to satisfy the constrained optimization problem. \cite{chow2017risk} propose conditional value-at-risk of accumulated cost and chance constraints as the values to be constrained and use a Lagrangian formulation to derive a Bellman optimality condition. \cite{dalal2018safe} use a different constraint for each MDP state and a safety layer that analytically solves a linearized action correction formulation per state. Similarly, \cite{pham2018optlayer} introduce a layer that corrects the output of a policy to respect constraints on the dynamic of a robotic arm.

\paragraph{Teacher Advice}
A subset of RL safety uses expert advice during exploration with  
potential-based reward shaping mechanisms~\citep{DBLP:conf/icml/NgHR99}.
\cite{wiewiora2003principled} introduce a general method for incorporating 
arbitrary advice into the reward structure. \cite{saunders2017trial} use a
human in the loop to learn an effective RL agent while minimizing cost
accumulated over training. \cite{DBLP:conf/socs/CamachoCSM17,
camacho2017decision} use DFAs with static reward shaping attached to states to
express non-Markovian rewards. We build on this with a learned reward shaping
function in the case of dense soft constraints, and by adding the translation
of MDP transitions into the symbols of the DFA alphabet. Similar to teacher
advice is shielding~\citep{jansen2018shielded,2018shieldedAAAI}, in which an agent's actions are filtered through a shield
that blocks actions that would introduce an unsafe state (similar to hard constraints; \autoref{formal_language_constraint_framework}).

\paragraph{Formal Languages} Formal languages and automata have been used before in RL for task specification or as task abstractions (options) in hierarchical reinforcement learning~\citep{DBLP:conf/icml/IcarteKVM18, DBLP:conf/iros/LiVB17, DBLP:conf/ijcai/WenPT17, DBLP:journals/corr/abs-1809-07823, DBLP:journals/asc/MousaviGMJ14}. In some cases, these automata were derived from Linear Temporal Logic (LTL) formulae, in others LTL or other formal language formulae have been directly used to specify tasks~\citep{DBLP:conf/atal/IcarteKVM18}. \cite{DBLP:journals/corr/LittmanTFIWM17} defines a modified LTL designed for use in reinforcement learning. In robotics, LTL is used for task learning~\citep{DBLP:conf/iros/LiVB17}, sometimes in conjunction with teacher demonstrations~\citep{DBLP:journals/corr/abs-1809-06305}.

\section{Formal Language Constrained MDPs}
\label{formal_language_constraint_framework}

The constrained Markov decision process (CMDP)~\citep{altman1999constrained} extends the Markov decision process~\citep{sutton2018reinforcement} to incorporate constraints. The difference is an additional set of cost functions $c_i: S \times A \times S \to \mathbb{R}$ and set of cost limits $d_i\in\mathbb{R}$. Then, the constrained optimization problem is
\begin{equation*}
\begin{aligned}
\argmax_{\pi} \quad & J_r(\pi)\\
\textrm{s.t.} \quad & J_{c_i}(\pi) \leq d_i, i=1,\ldots,k   \\
\end{aligned}
\end{equation*}
where $J_r(\pi)$ is a return-based objective, e.g., finite horizon discounted return defined $J_r(\pi) = \mathbb{E}_{\tau\sim\pi}[\sum_{t\in\tau} \gamma^t r_t]$ and $J_{c_i}$ is a cost-based constraint function defined similarly, replacing $r_t$ with $c_{i,t}$.

We propose formal language constrained MDPs (FLCMDPs) as a subset of CMDPs in
which each constraint $C_i\subset (S \times A \times S)^*$ is defined by a set
of prohibited trajectories. (Subscript \(i\) is suppressed from this point without loss of generality). Because $C$ is defined by a formal language, it can be
recognized efficiently by an automaton, which we use to construct
the cost function. We define three functions for interacting with the
constraint automaton: a translation function $T_C: (S \times A \times S) \to
\Sigma_C$ that converts MDP transitions into a symbol in the recognizer's input
language, a recognizer function $D_C: \Sigma_C \to Q_C$ that steps the recognizer
automaton using the input symbol and returns the state, and finally a cost assignment $G_C:
Q_C \to \mathbb{R}$ that assigns a real-valued cost to each recognizer state.
The composition of these three functions forms a CMDP cost function defined $c = G_C \circ
D_C \circ T_C: (S\times A\times S)\to\mathbb{R}$. The interaction of these functions with the underlying MDP framework is illustrated in Figure 1(a), where the constraint uses the MDP state and action at time $t$ to calculate the cost signal at time $t$ and, if action shaping is being employed as discussed below, influence the action at time $t+1$.

\paragraph{Translation Function $T_C$}
The translation function accepts as input the MDP state and action at each time step, and outputs a token in the discrete, finite language of the associated recognizer. This allows the recognizer automaton to be defined in a small, discrete language, rather than over unwieldy and potentially non-discrete MDP transitions. Further, freedom in choice of input language allows for flexible design of the constraint automaton to encode the desired inductive bias, and thus more meaningful structured states.

\paragraph{Recognizer Function $D_C$}
Each constraint is instantiated with a finite automaton recognizer that decides
whether a trajectory is in the constraint set. The only necessary
assumption about the recognizer is that it defines some meaningful state that
may be used for learning the constraint dynamics.  Our implementation uses a deterministic finite automaton (DFA)
as the recognizer for each constraint, defined as $(Q, \Sigma, \delta, q_0,
F)$, where $Q$ is the set of the DFA's states, $\Sigma$ is the alphabet over
which the constraint is defined, $\delta : Q \times A \to Q$ is the transition
function, $q_0 \in Q$ is the  start state, and $F \subset Q$ is the set of 
accepting states that represent constraint violations. The DFA is  set to its initial state at the start of each episode and is advanced at each time step with the token output by the translation layer. Although our experiments use DFAs as a relatively simple recognizer, the framework can be easily modified to work with automata that encode richer formal languages like pushdown automata or hybrid automata.

\paragraph{Constraint State Augmentation}
In order to more efficiently learn constraint dynamics, the MDP state $s_t$ is augmented with a one-hot representation of the recognizer state $q_t$. To preserve the Markov property of the underlying MDP,  state augmentation should contain sufficient information about the recognizer state and, if it is stateful, the translation function. To enhance performance, the one-hot state is embedded to $\floor{\log_2(|Q|)}$ dimensions before being input into any network and the embedding is learned with gradients backpropagated through the full network.

\paragraph{Cost Assignment Function}
The cost assignment function $G_C$ assigns a real-valued cost to each state of the recognizer. This cost can be used in optimization to enforce the constraint with a Lagrangian-derived objective penalty, or via reward shaping, which updates the reward function to $r_t - \lambda c_t$, where $\lambda$ is a scaling coeffficient.

Cost assignments are frequently sparse, where $G_C$ is only non-zero at accepting states that recognize a constraint violation. This poses a learning problem for optimization-based methods that  use reward shaping or an objective penalty to solve the CMDP. A goal of constrained RL is to minimize accumulated constraint violations over training but, to ensure that the frequency of violations is small, the optimization penalty can be large relative to the reward signal. This can lead to a situation in which an unnecessarily conservative policy is adopted early in training, slowing exploration. We next propose a method for learning a dense cost function that takes advantage of the structure of the constraint automaton to more quickly learn constraint dynamics and avoid unnecessarily conservative behavior.

\paragraph{Learned Dense Cost}
The goal of learning a dense cost is not to change the optimality or near-optimality of a policy with respect to the constrained learning problem. Thus, we use the form of potential-based shaping: $F(s_{t-1},a_t,s_{t}) = \gamma\Phi(s_t)-\Phi(s_{t-1})$,  where $\Phi$ is a potential function (see \cite{DBLP:conf/icml/NgHR99} for details). This is added as a shaping term to the sparse cost to get the dense cost
\[G'_C(q_{t-1}, q_{t}) = G_C(q_t) + \beta(\gamma\Phi(q_t) - \Phi(q_{t-1})) \enspace  , \]
where $\beta$ scales the added dense cost relative to the sparse cost, and  $\Phi$ is a function of the recognizer state rather than the MDP state, which requires $s_{t-2}$ and $a_{t-2}$ as additional inputs to calculate $q_{t-1}$. Generally, if the value of $\Phi$ increases as the automaton state is nearer to a violation, then the added shaping terms have the effect of adding cost for moving nearer to a constraint violation and refunding cost for backing away from a potential violation. 

In our experiments, the potential $\Phi^\pi(q_t)$ is defined using 
$t_v(q_t)$, which is 
a random variable defined as 
the number of steps between visiting  recognizer state $q_t$ and an accepting recognizer state. 
This variable's distribution is based on $\pi$ and the MDP's transition function. Its value  is small if a violation is expected to occur soon after reaching $q_t$ and vice-versa. We then define the potential function as 
$$\Phi^\pi(q_t) = \left( \frac{1}{2}\right)^{(\mathbb{E}_\pi[t_v(q_t)] / t_v^{baseline})} \enspace ,$$ 
which ensures that its value is always in $[0,1]$ and rises exponentially as the expected time to a violation becomes smaller. If the expected time to next violation is much larger than the provided baseline, $t_v^{baseline}$, then the potential value becomes small, as shaping is unnecessary in safe states. The expected value of $t_v(q_t)$ may be estimated empirically from rollouts, and is updated between episodes to ensure that it's stationary in each rollout. 
We set  $t_v^{baseline}$ to be the ratio of estimated or exact length of an episode and the constraint limit $d_i$, but find empirically that the the method is resilient to the exact choice.

\paragraph{Hard Constraints and Action Shaping}

When safety constraints are strict, i.e., when the limit on the number of constraint violations $d$ is zero, the set of available actions is reduced 
to ensure a violation cannot occur. If a constraint isn't fully state-dependent (i.e., there is always an action choice that avoids violation), then action shaping can guarantee that a constraint is never violated. Otherwise, knowledge of which actions lead to future violating trajectories requires knowledge of the underlying MDP dynamics, which is possible by learning a model that converts state constraints into state-conditional action constraints as in~\cite{dalal2018safe}.

Our implementation of hard constraints initially allows the agent to freely choose its action, but before finalizing that choice, simulates stepping the DFA with the resulting token from the translation function and, if that lookahead step would move it into a violating state, it switches to the next best choice until a non-violating action is found. For the constraints in our experiments, it is always possible to choose a non-violating action. A known safe fallback policy can be employed in the case when an episode cannot be terminated. Action shaping can be applied during training or deployment, as opposed to reward shaping,  which is only applied during training. Thus, we experiment with applying action shaping only during training, only during evaluation, or in both training and evaluation.

\section{Experimental Evaluation}
\label{experiments}

\subsection{Constraints}

We evaluated FLCMDPs on four families of constraints, which we define with regular expressions.


{\bf No-dithering:}
 A no-dithering constraint prohibits movements in small, tight patterns that cover very small areas. In one dimension, we define dithering as actions are taken to move left, right, left, and right in order or the opposite, i.e.,
$.^* \, (\ell \, r)^2|(r \, \ell)^2$.
The automaton encoding this constraint is depicted in
\ifMergeFigs
Figure~\ref{fig:merged}(b).
\else
Figure~\ref{fig:nodither}.
\fi
In environments with two-dimensional action spaces, such as Atari Seaquest, we generalize this to vertical and diagonal moves and constrains actions that take the agent back to where it started in at most four steps\footnote{The regex describing this constraint is included in Appendix~\ref{app:no_dithering_2d_regex}.}.
In MuJoCo, constraints are applied per joint and the translation function maps negative and positive-valued actions to `$\ell$' and `$r$', respectively.

\textbf{No-overactuating:}
A no-overactuating constraint prohibits repeated movements in the same direction over a long period of time. In Atari environments, this forbids moving in the same direction four times in a row, i.e.,
$.^* \, (\ell^4 \cup r^4)$.
In two dimensions, this is extended to include moving vertically: $.^* \, (L^4 \cup R^4 \cup U^4 \cup D^4)$. Each of the left ($L$), right ($R$), up ($U$) and down ($D$) tokens is produced by the translation function from the primary direction it's named after or diagonal moves that contain the primary direction, e.g., $L = \ell  \cup \ell$+$ u \cup \ell$+$d$, where ``$\ell$+$ u$'' is the atomic left-up diagonal action. In MuJoCo environments, overactuation is modelled as occurring when the sum of the magnitudes of joint actuations exceeds a threshold. This requires the translation function to discretize the magnitude in order for a DFA to calculate an approximate sum. The MDP state-based version is ``dynamic actuation'', which sets the threshold dynamically based on a discretized distance from the goal.

\textbf{Proximity:} The proximity constraint, used in Safety Gym, encodes the distance to a collision with any of the variety of hazards found in its environments. The translation function uses the max value over all the hazard lidars, which have higher value as a hazard comes closer, and discretizes it into one of ten values. The constraint is defined as being violated if the agent contacts the hazard, which is identical to the constraint defined in the Safety Gym environments and described in~\cite{ray2019benchmarking}.

\textbf{Domain-specific:}
In addition to the previously described simple constraints, we define hand-built constraints for the Breakout and Space Invaders Atari environments.  These constraints are designed to mimic specific human strategies in each environment for avoiding taking actions that end the episode.  In Atari Breakout, we define the ``paddle-ball'' constraint, which limits the allowed horizontal distance between the ball and the center of the paddle. In Atari Space Invaders, we define the ``danger zone'' constraint, which puts a floor on the the allowed distance between the player's ship and the bullets fired by enemies. We provide more details of each constraint in Appendix~\ref{app:constr-transl-funct}.

\begin{table*}[t]
    \centering
    \caption{Metrics averaged over the last 25 episodes of training in Safety Gym environments with PPO-Lagrangian methods, normalized relative to unconstrained PPO metrics. Cost rate is the accumulated cost regret over the entirety of training.
    }\label{tab:safety-gym-lagrangian}
    \vspace{1mm}
    \input{safety-gym-lagrangian}
    \vspace*{-0.125in}
\label{tab:atari_soft_mincost}
\end{table*}

\begin{figure}
    \centering
    \includegraphics[width=.38\textwidth]{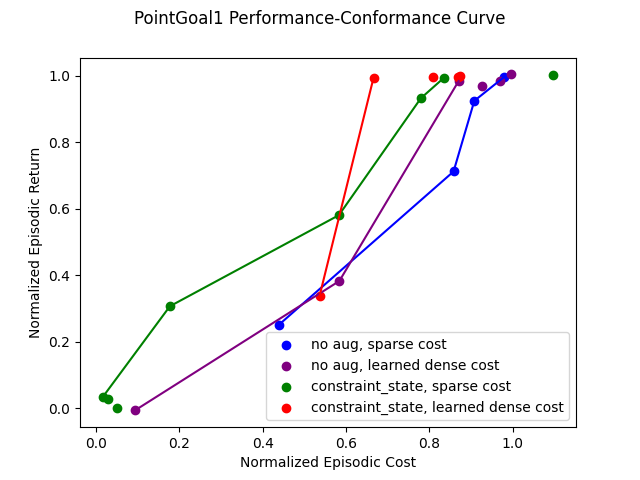}
    \includegraphics[width=.38\textwidth]{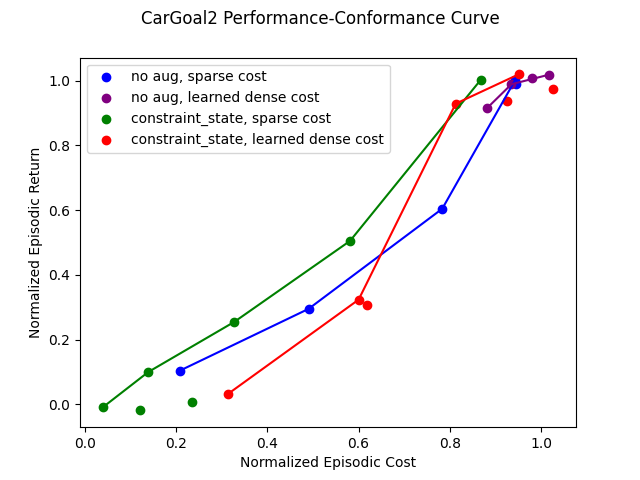}
    \includegraphics[width=.38\textwidth]{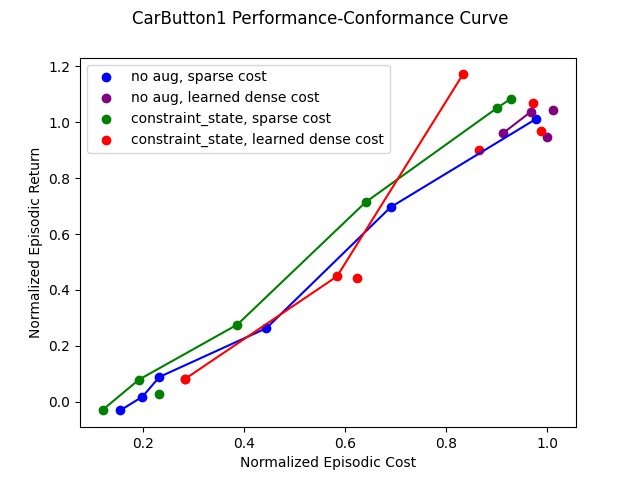}
    \includegraphics[width=.38\textwidth]{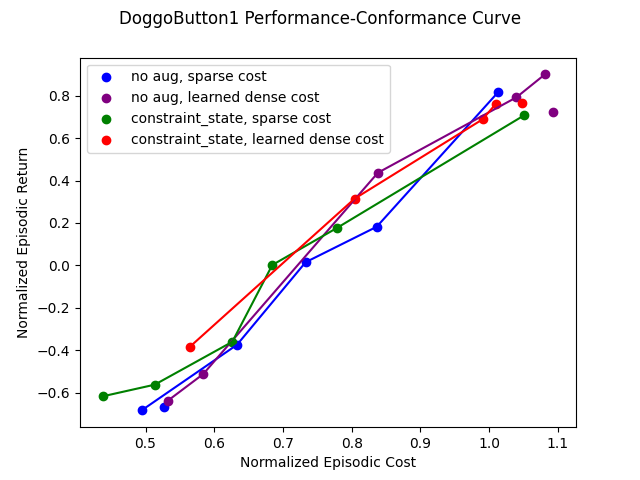}
    \caption{Performance/conformance curves in selected Safety Gym environments with Pareto frontiers plotted per reward shaping method. We observe that using state augmentation (green) consistently outperforms the baseline (blue) at all levels of reward shaping, which are anti-correlated with episodic cost and episodic return. The use of cost shaping (purple) produces gains in return at a given amount of cost only at small reward shaping values correlating to high return and cost. Consequently, the combination of state augmentation and cost shaping inherits this behavior of being more effectiveness when cost/return are higher. The full set of plots is included in Appendix~\ref{app:safety-gym-reward-shaping-plots}.}
    \label{fig:my_label}
\end{figure}

\subsection{Environments}
\label{sec:environments_and_experiments}

In Safety Gym, the Spinning Up implementation of PPO with Lagrangian optimization penalization was employed, with hyperparameters as chosen identically to~\cite{ray2019benchmarking}. We modified each network to concatenate the constraint state augmentation with the input and used $d = 25$ for the expected cost limit. All safety requirements are accounted for in a single constraint and we report the constraint violations as accounted for in the Safety Gym environments rather than as reported by the finite automaton (though these are identical when not using cost shaping). Each environment, which is randomly re-arranged each episode, is made up of a pairing of a robot and a task. The robots are Point, which turns and moves, and Car, which is wheeled with differential drive control. The tasks are Goal, which requires moving into a goal area, Button, which requires pressing a series of buttons, and Push, which requires moving a box into a goal area. More details can be found in~\cite{ray2019benchmarking}.

In Atari environments~\citep{bellemare13arcade}, we modified the DQN implemented in OpenAI Baselines~\citep{baselines} by appending the state augmentation to the output of its final convolutional layer. Reward shaping was used for soft constraint enforcement with the penalty fixed at one of $\{0, -0.001,-0.0025,-0.005,-0.01\}$, and each agent was trained for 10M steps before collecting data from an evaluation phase of 100K steps for 15 or more train/eval seed pairs for each hyperparameter combination.
For MuJoCo environments~\citep{brockman2016openai}, we trained the Baselines PPO agent~\citep{schulman2017proximal} with dense constraints and the state augmentation concatenated to the input MDP observation. Reward shaping similar to Atari was employed for constraint enforcement.
Atari and MuJoCo environments do not have constraints built in, so we report the number of constraint violations per episode from the custom constraints and minimize them without a specific goal value.

\subsection{Results}

We experiment with our methods to evaluate the usefulness of formal language constraints in optimizing three objectives. In the final policy output by the training process, it is desirable to simultaneously maximize return per episode and minimize constraint violations per episode, or keep them below the specified limit. The third objective is to minimize accumulated cost regret over the course of training. To examine the proposed methods, we investigate two questions. First, what effect do the proposed methods have on accumulated regret? In Section~\ref{subsec:lagrangian}, we compare the proposed methods against a baseline when combined with PPO using a Lagrangian approach in Safety Gym~\citep{ray2019benchmarking}.
Second, how should hyperparameters be chosen to minimize or maximize each objective respectively? In Section~\ref{subsec:reward_shaping} we examine which hyperparameter choices worked well in various Atari and MuJoCo environments using reward shaping. Finally, in Section~\ref{subsec:hard_shaping} we see what effect enforcing zero constraint violations with action shaping has on Atari environments. 

\subsubsection{Lagrangian results and accumulated cost}
\label{subsec:lagrangian}

Table~\ref{tab:safety-gym-lagrangian} compares PPO with Lagrangian constraint enforcement with and without constraint state augmentation. The clearest trend is in the reduction of the cost rate, which measures accumulated cost regret, often by between almost one half and an order of magnitude. This results from the inclusion of the helpful inductive bias provided by the constraint structure. This result is not surprising, but does quantify the magnitude of the benefit that a low-overhead method like formal language constraints can have. Qualitatively, we noted that the earliest steps of training had decreased performance generally as the embedding of the constraint state was being learned, but quickly surpassed baseline performance once the updates of the embedded representation became small.

\cite{ray2019benchmarking} says that algorithm $A_1$ dominates
$A_2$ when they are evaluated under the same
conditions, the cost and return of $A_1$ are at least as good as that of $A_2$, and at least one of cost and return is strictly better for $A_1$. By this definition, the state augmented approach strictly dominates the baseline in 6 of 12 environments, while coming close in most of the rest. Specifically, we also note that state augmentation allowed a significant step to be taken in closing the gap between unconstrained PPO return and PPO-Lagrangian in the Point-Goal2 and Car-Push2 environments, with increases of roughly an order of magnitude in each.

\begin{table*}[t]
    \caption{Atari reward shaping with state augmentation, choosing hyperparameters that minimize constraint violations per episode. ``Dense'' refers to whether the dense cost term was used and ``reward shaping'' refers to the fixed reward shaping coefficient $\lambda$.}\label{param-select-min-viols}
    \resizebox{\textwidth}{!}{\input{min-violations-env-constraint}}
    \vspace*{-0.125in}
\label{tab:atari_soft_mincost}
\end{table*}

\subsubsection{Reward shaping results and sensitivity to reward shaping}
\label{subsec:reward_shaping}

\begin{table*}[t]
    \caption{Atari reward shaping with state augmentation, choosing hyperparameters that maximize cumulative reward per episode. ``Dense'' refers to whether the dense cost term was used and ``reward shaping'' refers to the fixed reward shaping coefficient $\lambda$.}\label{param-select-max-reward}
    \resizebox{\textwidth}{!}{\input{max-reward-environment-constraint}}
\vspace*{-0.125in}
\label{tab:atari_soft_maxreward}
\end{table*}

The most basic function of the proposed framework is to reduce constraint violations. Table~\ref{tab:atari_soft_mincost} presents the mean and standard deviation of violations per 100 evaluation steps, episode length, and episode reward for reward shaping-enforced constraints with the choice of hyperparameters that produced the minimum number of violations for each environment/constraint pair in evaluation. We note that the highest value of reward shaping available is generally the best choice for minimizing constraint violations, which were often reduced by an order of magnitude or more from the baseline. Minimizing constraint violations has a small deleterious effect on mean episode reward, but because mean reward per step didn't decrease, this implies that episodes were shorter as a result of constraint enforcement.

In addition to minimizing constraint violations, we found that the application of soft constraints can also increase reward per episode. Table~\ref{param-select-max-reward} presents results for soft constraints with the choice of hyperparameters that produced the maximum reward in each environment/constraint pair. In this case, lower reward shaping values perform best. The hyperparameter values that minimized constraint violations with the Breakout actuation and paddle ball constraints also maximized reward, implying that the objectives were correlated under those constraints. Table~\ref{tab:mujoco-soft} presents results for soft constraints with constraint state augmentation in three MuJoCo environments. We find, similar to Atari, that there is one value of reward shaping that is most effective in each environment/constraint pair and that reward degrades smoothly as is shifted from the optimal value.

\begin{table*}[t]
\caption{Mean per-episode MuJoCo rewards and violations with soft dense constraints and constraint state augmentation. Top row displays the reward shaping coefficient $\lambda$.} \label{table_mujoco_ss}
  \centering
  \resizebox{\textwidth}{!}{\input{mujoco-results}}
  \label{tab:mujoco-soft}
\end{table*}

\begin{table*}[t]
    \caption{Atari results with hard constraints, choosing hyperparameters which maximize reward when applying action shaping in training and evaluation, only in training, or only in evaluation.}\label{hard-results}
    \centering
    \resizebox{\textwidth}{!}{\input{hard-results}}
\vspace*{-0.125in}
\end{table*}

\subsubsection{Hard Action Shaping Results}
\label{subsec:hard_shaping}

Table~\ref{hard-results} presents results for hard constraints with the hyperparameters that produced the maximum return for each environment/constraint pair. Results for cases where hard action shaping was only applied during training or only applied during evaluation are presented as well. There is a slight trend indicating that using action shaping at train time in addition to evaluation increases performance. For those constraints that are qualitatively observed to constrain adaptive behavior, performance rises when using hard shaping only in training, at the cost of allowing constraint violations.

\section{Discussion}
\label{sec:discussion}

The ability to specify MDP constraints in formal languages opens the possibility for using model checking~\citep{kupferman2000automata,bouajjani1997reachability}, agnostic to the choice of learning algorithm, to verify  properties of a safety constraint. 
Formal language
constraints might be learned from exploration, given a pre-specified safety objective, and, because of their explicitness, used without complication for downstream applications or verification.
This makes 
formal language
constraints particularly 
useful in multi-component, contract-based software systems~\citep{meyer1992applying}, where one or more components is learned using the MDP formalism.

Experiments with more complex constraints are necessary to explore yet unaddressed challenges,
the primary challenge being 
that the constraints used with action shaping in this work were all ``best effort'', i.e., the allowed set of actions was never be empty. If this is not the case, lookahead might be required to guarantee zero constraint violations.
Further, the tested hard constraints were only used with DQN, which provides a ranked choice over discrete actions. Future work might investigate how to choose optimal actions which are not the first choice in the absence of ranked choice or an explicit fallback policy.

\newpage

\bibliography{iclr2021_conference}
\bibliographystyle{iclr2021_conference}

\newpage
\appendix

\section{Agent and Environment Details}
\label{app:agent-envir-deta}

\subsection{Atari}
\label{sec:atari}

We used the OpenAI Baselines implementation of DQN with default settings with all Atari environments. The only modification to the model was the concatenation of any state augmentation (where employed) with the output of the convolutional layers. From a previous set of experiments, we found that the use of prioritized replay dramatically shrinks the optimal values of the reward shaping coefficient, but that the optimal choice is still strongly based on the choice of environment. We hypothesize that this is largely due to the differing MDP dynamics and reward frequency.

\subsection{MuJoCo}
\label{sec:mujoco}

We used the unmodified OpenAI baselines implementation of PPO in all MuJoCo environments, simply appending any state augmentation to the MDP observation as input to the model.

\section{Constraints and Translation Functions}
\label{app:constr-transl-funct}

\subsection{Atari Breakout Paddle Ball Distance}
\label{sec:atari:-breakout-1}

We designed the ``paddle ball distance'' constraint for the Breakout environment. The translation function uses the last frame from each environment observation and calculates, using rudimentary computer vision provided by scikit-image~\cite{scikit-image}, the horizontal distance between the center of the ball and the center of the paddle. Then, if the paddle was too far to the left relative to the ball and the ``move right'' action wasn't taken that frame, the ``L'' token is output and similar for the ``R'' token. Otherwise, if the ball is outside the area under the bricks or no other token is output, a zero token is output.

The constraint is a simple counter that  increments an L counter for each successive L token and similar for R and accepts on 3 successive, identical tokens. Receiving a zero token or a token from the other direction resets each counter.

We chose 10 pixels as the maximum allowed distance between ball and paddle under this constraint, but this turned out to be too restrictive for good performance under hard constraints.

\subsection{Atari: Space Invaders}
\label{sec:atar-space-invad-1}

The translation function for ``dangerzone'' uses scikit-learn to find the position of each downward-travelling bullet fired by enemies in the area immediately above the player's ship, as well as the position of the player's ship. Then, it calculates the distance between the closest bullet and the ship, as well as whether the bullet is to the right, left, or above the ship. Each token is a concatenation of the direction of the bullet (left, right, or above) and a discretization of the distance ($x \le 12$ pixels, $12 < x \le 24$ pixels, or $x >24$ pixels).

The constraint is violated if a bullet is at the closest distance, $<12$ pixels, and an action to dodge isn't taken, i.e., moving right from a bullet to the left, moving left from a bullet on the right, or moving in either direction from a bullet above.

\subsection{MuJoCo: Reacher}
\label{sec:mujoco:-reacher-1}

Two related constraints were explored in the Reacher environment: an actuation constraint similar to those presented in Atari, and a dynamic actuation constraint that  was modified to take the relative position of the ball to the goal into account. The translation function each used discretizes the action in increments of $0.2$. Each used a DFA to track the sum of actuation values which accepts, if the sum over 3 timesteps is greater than a threshold of $4.0$. In the dynamic reacher constraint, the acceptance threshold varies by how close the goal is.

\subsection{MuJoCo: Half Cheetah}
\label{sec:mujoco:-half-cheetah}

The Half Cheetah environment uses the one dimensional no-dithering constraint (similar to Atari) on each of the six joints of the simulated robot. The translation function discretizes positive and negative joint forces to be right and left, respectively. 

\section{Safety Gym Reward Shaping Plots}
\label{app:safety-gym-reward-shaping-plots}

\begin{figure}[h!]
  \centering
  \includegraphics[width=.45\textwidth]{perfconf_plots/CarButton1_performance_conformance.png}
    \includegraphics[width=.45\textwidth]{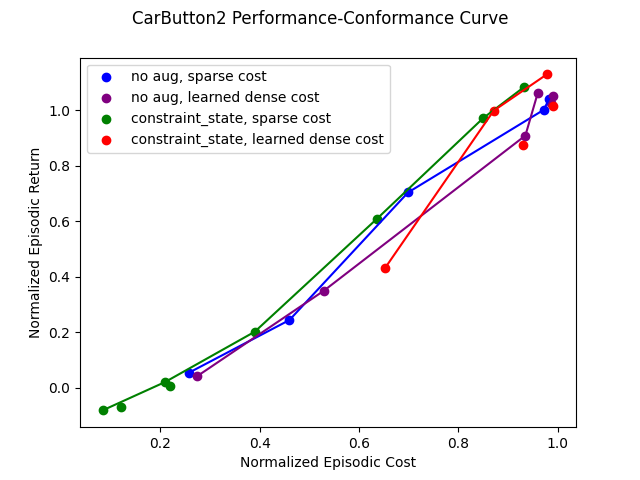}
  \caption{Performance/Conformance curves for CarButton environments,
    with Pareto frontiers plotter per reward shaping method.}
  \label{fig:carbutton1_performance_conformance}
\end{figure}

\begin{figure}[h!]
  \centering
  \includegraphics[width=.45\textwidth]{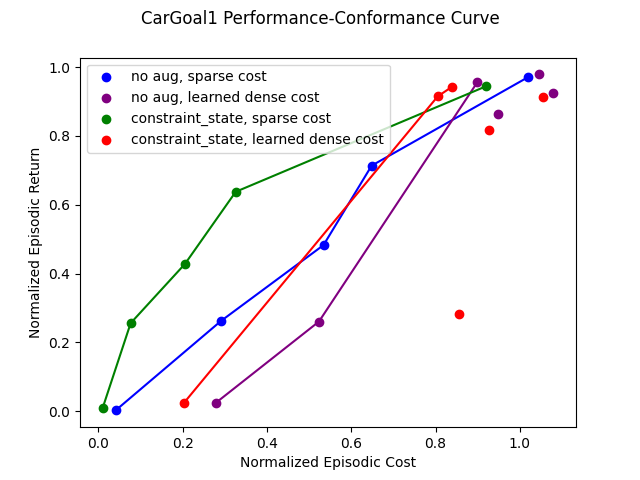}
  \includegraphics[width=.45\textwidth]{perfconf_plots/CarGoal2_performance_conformance.png}
  \caption{Performance/Conformance curves for CarGoal environments,
    with Pareto frontiers plotted per reward shaping method.}
  \label{fig:cargoal1_performance_conformance}
\end{figure}

\begin{figure}[h!]
  \centering
  \includegraphics[width=.45\textwidth]{perfconf_plots/DoggoButton1_performance_conformance.png}
  \includegraphics[width=.45\textwidth]{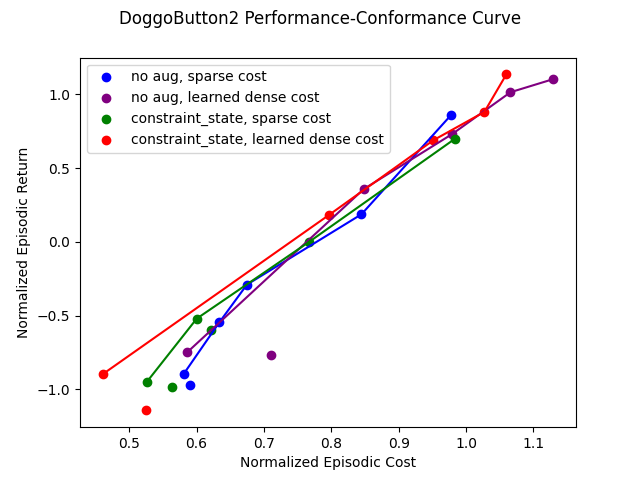}
  \caption{Performance/Conformance curves for DoggoButton environments,
    with Pareto frontiers plotted per reward shaping method.}
  \label{fig:doggobutton1_performance_conformance}
\end{figure}

\begin{figure}[h!]
  \centering
  \includegraphics[width=.45\textwidth]{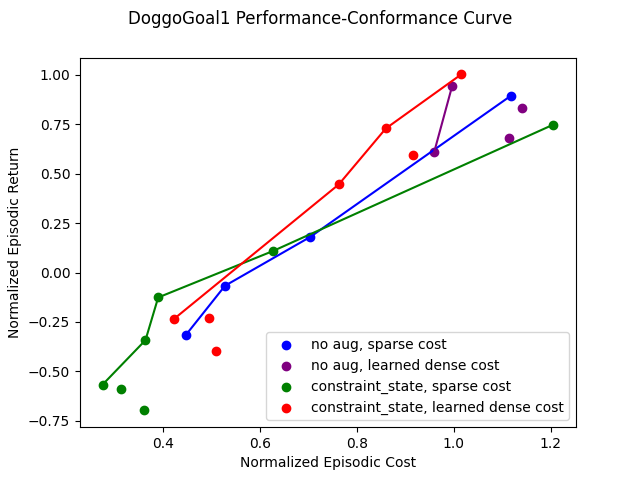}
  \includegraphics[width=.45\textwidth]{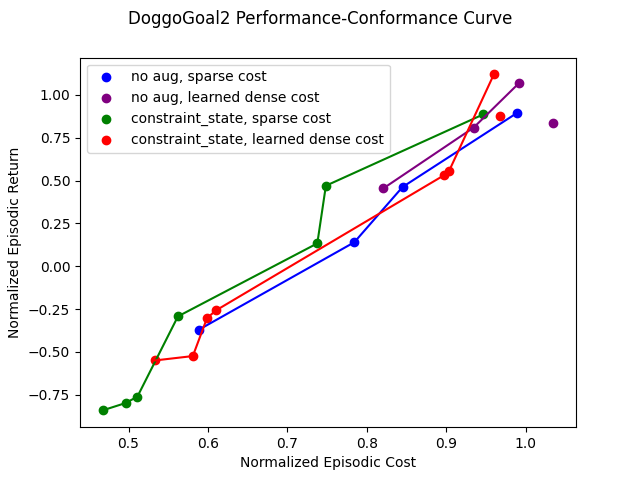}
  \caption{Performance/Conformance curves for DoggoGoal environments,
    with Pareto frontiers plotted per reward shaping method.}
  \label{fig:doggogoal1_performance_conformance}
\end{figure}

\begin{figure}[h!]
  \centering
  \includegraphics[width=.45\textwidth]{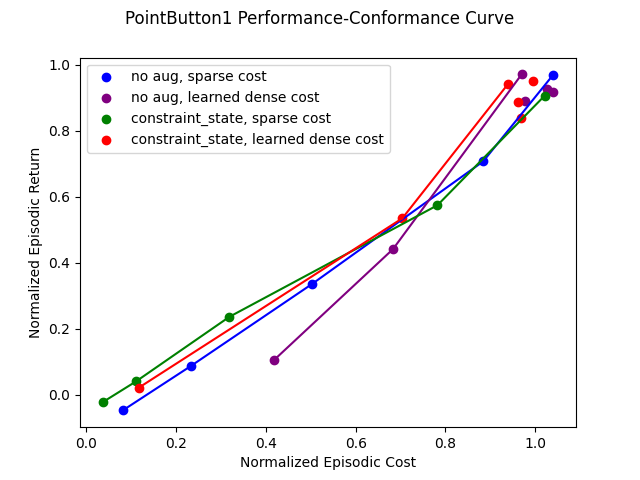}
  \includegraphics[width=.45\textwidth]{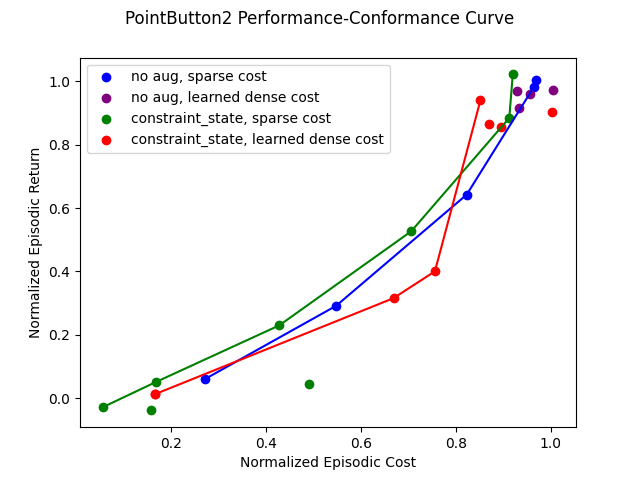}
  \caption{Performance/Conformance curves for PointButton environments,
    with Pareto frontiers plotted per reward shaping method.}
  \label{fig:pointbutton1_performance_conformance}
\end{figure}

\begin{figure}[h!]
  \centering
  \includegraphics[width=.45\textwidth]{perfconf_plots/PointGoal1_performance_conformance.png}
  \includegraphics[width=.45\textwidth]{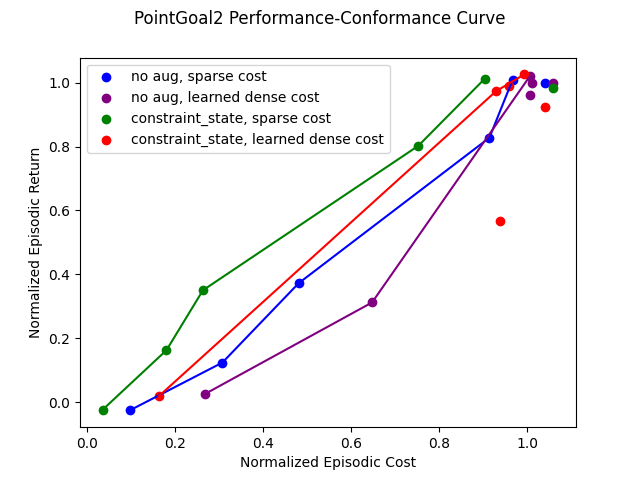}
  \caption{Performance/Conformance curves for PointGoal environments,
    with Pareto frontiers plotted per reward shaping method.}
  \label{fig:pointgoal1_performance_conformance}
\end{figure}

\section{Full 2D No-Dithering Regex}
\label{app:no_dithering_2d_regex}

\input{seaquest_2d_regex}

\end{document}

%% file: math_commands.tex
\usepackage{amsmath,amsfonts,bm}

\def\eqref#1{equation~\ref{#1}}

\def\floor#1{\lfloor #1 \rfloor}
\def\1{\bm{1}}

\def\rv{{\textnormal{v}}}

\DeclareMathAlphabet{\mathsfit}{\encodingdefault}{\sfdefault}{m}{sl}
\SetMathAlphabet{\mathsfit}{bold}{\encodingdefault}{\sfdefault}{bx}{n}

\DeclareMathOperator*{\argmax}{arg\,max}

%% file: flc-op-through-time.tex
\begin{tikzpicture}[node distance=4cm and 4cm]
  \tikzset{
    boxy/.style={
      rectangle,
      rounded corners,
      draw=black,
      minimum height=4em,
      minimum width=8em,
      align=center
    },
    ovaly/.style={
      ellipse,
      draw=black,
      align=center
    }
  }
  \node[boxy] (mdp1)  {MDP};
  \node[boxy] (mdp2) [right of=mdp1] {MDP};
  \node[ovaly] (tl1) [below = 1.5cm of mdp1] {Translation \\ Function $T_C$};
  \node[ovaly] (tl2) [below = 1.5cm of mdp2] {Translation \\ Function $T_C$};
  \node[boxy] (recog1) [below = 1.5cm of tl1] {Recognizer $D_C$};
  \node[boxy] (recog2)[below = 1.5cm of tl2] {Recognizer $D_C$};
  \node (arrowl1) [left=.75cm of mdp1] {};
  \draw[->] (arrowl1) edge (mdp1);
  \node (arrowl2) [left=.75cm of tl1] {};
  \draw[->] (arrowl2) edge (tl1);
  \node (arrowl3) [left=.75cm of recog1] {};
  \draw[->] (arrowl3) edge (recog1);
  \node (arrowr1) [right=1cm of mdp2] {};
  \draw[->] (mdp2) edge (arrowr1);
  \node (arrowr2) [right=1cm of tl2] {};
  \draw[->] (tl2) edge (arrowr2);
  \node (arrowr3) [right=1cm of recog2] {};
  \draw[->] (recog2) edge (arrowr3);
  \draw[->] (mdp1) edge node[above]{state} (mdp2);
  \draw[->] (recog1) edge node[above]{state} (recog2);
  \draw[->] (tl1) edge node[left]{token} (recog1);
  \draw[->] (tl2) edge node[left]{token} (recog2);
  \draw[->] (mdp1) edge node{\(a_t\), \(s_t\)} (tl1);
  \draw[->] (mdp2) edge node{\(a_{t + 1}\), \(s_{t + 1}\)} (tl2);
  \draw[dotted,->] (recog1) edge node[below,align=center]{Action\\Shaping} (mdp2);
  \draw[dotted,->] (recog2) edge[bend right=45] node[below=1cm]{Cost $Q_C$} (mdp2);
\end{tikzpicture}


%% file: no-1d-dithering-dfa.tex
\begin{tikzpicture}[->,node distance=1.7cm,scale=0.15]
  \tikzstyle{every state}=[draw]
  \node[initial,state] (I)              {$q_0$};
  \node[state]         (L) [above right of=I] {$q_1$};
  \node[state]         (LR) [right of=L] {$q_2$};
  \node[state]         (LRL) [right of=LR] {$q_3$};
  \node[state,accepting]         (V) [below right of=LRL] {$q_v$};
  \node[state]         (R) [below right of=I] {$q_4$};
  \node[state]         (RL) [right of=R] {$q_5$};
  \node[state]         (RLR) [right of=RL] {$q_6$};
  \path (I)   edge [loop above] node [above] {$n$,$f$} (I)
              edge              node [above] {$\ell$} (L)
              edge              node [below] {\(r\)} (R)
        (L)   edge              node [below] {$r$} (LR)
        (LR)  edge              node [below] {$\ell$} (LRL)
        (LRL) edge              node [right] {$r$} (V)
        (R)   edge              node [below] {\(\ell\)} (RL)
        (RL)  edge              node [below] {\(r\)} (RLR)
        (RLR) edge              node [right] {\(\ell\)} (V);
\end{tikzpicture}

%% file: safety-gym-lagrangian.tex
\begin{tabular}{ccccccc}
  \toprule
  & \multicolumn{3}{c}{FLCMDP State Augmented} & \multicolumn{3}{c}{Baseline} \\
  \cmidrule(lr){2-4}\cmidrule(lr){5-7}
 Environment & Return  & Violation & Cost Rate & Return & Violation & Cost Rate \\
  \midrule
 Point-Goal1 & 0.750 & \textbf{0.427} & \textbf{0.281} & \textbf{0.918} & 0.925 & 0.503 \\
 Point-Goal2 & \textbf{0.195} & 0.083 & \textbf{0.078} & 0.021 & \textbf{0.062} & 0.155 \\
 Point-Button1 & 0.252 & \textbf{0.129} & \textbf{0.128} & \textbf{0.343} & 0.296 & 0.218 \\
 Point-Button2 & \textbf{0.251} & \textbf{0.130} & 0.141 & 0.166 & 0.255 & \textbf{0.118} \\
 Point-Push1 & 0.549 & \textbf{0.042} & \textbf{0.061} & \textbf{0.692} & 0.496 & 0.543 \\
  Point-Push2 & \textbf{0.938} & \textbf{0.173} & \textbf{0.148} & 0.670 & 0.295 & 0.258 \\
  Car-Goal1 & \textbf{0.825} & \textbf{0.295} & \textbf{0.284} & 0.803 & 0.475 & 0.445 \\
 Car-Goal2 & 0.005 & \textbf{0.011} & \textbf{0.079} & \textbf{0.021} & 0.046 & 0.108 \\
 Car-Button1 & \textbf{0.022} & 0.083 & \textbf{0.071} & 0.018 & \textbf{0.039} & 0.118 \\
  Car-Button2 & \textbf{0.031} & 0.147 & \textbf{0.076} & 0.009 & \textbf{0.009} & 0.078 \\
 Car-Push1 & 0.737 & \textbf{0.032} & \textbf{0.069} & \textbf{0.882} & 0.387 & 0.420 \\
  Car-Push2 & \textbf{0.256} & \textbf{0.086} & \textbf{0.124} & 0.025 & 0.115 & 0.202 \\
 \bottomrule
\end{tabular}

%% file: min-violations-env-constraint.tex
\begin{tabular}{lllccccccc}
  \toprule
   &  &  &  & \multicolumn{3}{c}{\textbf{FLCMDP State Augmented}} & \multicolumn{3}{c}{\textbf{Baseline}}                                                   \\
  \cmidrule(lr){5-7} \cmidrule(lr){8-10}
                               &                     &                & \textbf{Reward}  & \textbf{Mean}                 & \textbf{Mean}                & \textbf{Mean}                   & \textbf{Mean}                   & \textbf{Mean}              & \textbf{Mean}            \\
\textbf{Environment}           & \textbf{Constraint} & \textbf{Dense} & \textbf{Shaping} & \textbf{Episode Reward}       & \textbf{Step Reward}         & \textbf{Viols/100 steps}        & \textbf{Episode Reward}         & \textbf{Step Reward}       & \textbf{Viols/100 steps} \\
\midrule                                                                                                                                                                                                                                                           
\multirow{3}{*}{Breakout}      & actuation           & False          & \(-0.001\)       & \textbf{297.12 \(\pm\) 8.07}  & \textbf{0.15 \(\pm\) 0.0039} & \textbf{0.45 \(\pm\) 0.00067}   & 272.19 \(\pm\) 43.12            & 0.14 \(\pm\) 0.01          & 13.59 \(\pm\) 0.025      \\
                               & dithering           & False          & \(-0.001\)       & 263.57 \(\pm\) 11.14          & \textbf{0.15 \(\pm\) 0.0068} & \textbf{\(0.0008 \pm 8.4e-06\)} & \textbf{272.19 \(\pm\) 43.12}   & 0.14 \(\pm\) 0.01          & 0.12 \(\pm\) 0.001       \\
                               & paddle ball         & True           & \(-0.0025\)      & \textbf{314.79 \(\pm\) 15.09} & \textbf{0.17 \(\pm\) 0.0056} & \textbf{6.15 \(\pm\) 0.0031}    & 272.19 \(\pm\) 43.12            & 0.14 \(\pm\) 0.01          & 13.40 \(\pm\) 0.022      \\
  \midrule                                                                                                                                                                                                                                                         
\multirow{2}{*}{Seaquest}      & actuation           & False          & \(-0.01\)        & 1858.65 \(\pm\) 478.56        & 0.76 \(\pm\) 0.18            & \textbf{2.71 \(\pm\) 0.0066}    & \textbf{2250.13 \(\pm\) 647.92} & \textbf{0.96 \(\pm\) 0.21} & 9.74 \(\pm\) 0.017       \\
                               & dithering           & False          & \(-0.01\)        & 1608.66 \(\pm\) 41.25         & 0.75 \(\pm\) 0.013           & \textbf{0.081 \(\pm\) 0.001}    & \textbf{2250.13 \(\pm\) 647.92} & \textbf{0.96 \(\pm\) 0.21} & 1.61 \(\pm\) 0.007       \\
  \midrule                                                                                                                                                                                                                                                         
\multirow{3}{*}{SpaceInvaders} & actuation           & False          & \(-0.01\)        & 598.78 \(\pm\) 39.98          & \textbf{0.63 \(\pm\) 0.033}  & \textbf{5.39 \(\pm\) 0.017}     & \textbf{604.86 \(\pm\) 44.86}   & 0.62 \(\pm\) 0.04          & 10.88 \(\pm\) 0.011      \\
                               & dangerzone          & True           & \(-0.005\)       & \textbf{629.32 \(\pm\) 28.72} & \textbf{0.65 \(\pm\) 0.017}  & 0.00 \(\pm\) 0.00               & \textbf{604.86 \(\pm\) 44.86}   & 0.62 \(\pm\) 0.04          & 0.00 \(\pm\) 0.00        \\
                               & dithering           & True           & \(-0.01\)        & 595.25 \(\pm\) 20.25          & \textbf{0.63 \(\pm\) 0.021}  & \textbf{0.00 \(\pm\) 0.00}      & \textbf{604.86 \(\pm\) 44.86}   & 0.62 \(\pm\) 0.04          & 0.53 \(\pm\) 0.0064      \\
\bottomrule
\end{tabular}

%% file: max-reward-environment-constraint.tex
\begin{tabular}{lllccccccc}
  \toprule
  & & & & \multicolumn{3}{c}{\textbf{FLCMDP State Augmented}} & \multicolumn{3}{c}{\textbf{Baseline}}                                                       \\
  \cmidrule(lr){5-7} \cmidrule(lr){8-10}
                               &                     &                &  \textbf{Reward} & \textbf{Mean}                 & \textbf{Mean}                & \textbf{Mean}                 & \textbf{Mean}                   & \textbf{Mean}              & \textbf{Mean}                \\ 
\textbf{Environment}           & \textbf{Constraint} & \textbf{Dense} & \textbf{Shaping} & \textbf{Episode Reward}       & \textbf{Step Reward}         & \textbf{Viols/100 Steps}      & \textbf{Episode Reward}         & \textbf{Step Reward}       & \textbf{Viols/100 Steps}     \\
\midrule                                                                                                                                                                                                                                                           
\multirow{3}{*}{Breakout}      & actuation           & False          & \(-0.001\)       & \textbf{297.12 \(\pm\) 8.07}  & \textbf{0.15 \(\pm\) 0.0039} & \textbf{0.45 \(\pm\) 0.00067} & 272.19 \(\pm\) 43.12            & 0.14 \(\pm\) 0.01          & 13.59 \(\pm\) 0.025          \\
                               & dithering           & True           & \(-0.005\)       & \textbf{302.24 \(\pm\) 43.81} & 0.14 \(\pm\) 0.02            & \textbf{0.11 \(\pm\) 0.001}   & 272.19 \(\pm\) 43.12            & 0.14 \(\pm\) 0.01          & 0.12 \(\pm\) 0.001           \\
                               & paddle ball         & True           & \(-0.0025\)      & \textbf{314.79 \(\pm\) 15.09} & \textbf{0.17 \(\pm\) 0.0056} & \textbf{6.15 \(\pm\) 0.0031}  & 272.19 \(\pm\) 43.12            & 0.14 \(\pm\) 0.01          & 13.40 \(\pm\) 0.022          \\
\midrule                                                                                                                                                                                                                                                           
\multirow{2}{*}{Seaquest}      & actuation           & True           & \(-0.0025\)      & 2339.54 \(\pm\) 442.02        & 0.93 \(\pm\) 0.11            & \textbf{4.43 \(\pm\) 0.016}   & \textbf{2250.13 \(\pm\) 647.92} & \textbf{0.96 \(\pm\) 0.21} & 9.74 \(\pm\) 0.017           \\
                               & dithering           & False          & \(-0.001\)       & 1997.91 \(\pm\) 539.75        & 0.86 \(\pm\) 0.23            & \textbf{1.58 \(\pm\) 0.011}   & \textbf{2250.13 \(\pm\) 647.92} & \textbf{0.96 \(\pm\) 0.21} & 1.61 \(\pm\) 0.007           \\
\midrule                                                                                                                                                                                                                                                           
\multirow{3}{*}{SpaceInvaders} & actuation           & False          & \(-0.005\)       & \textbf{646.99 \(\pm\) 50.55} & \textbf{0.64 \(\pm\) 0.04}   & 29 \(\pm\) 0.0053             & 604.86 \(\pm\) 44.86            & 0.62 \(\pm\) 0.04          & \textbf{10.88 \(\pm\) 0.011} \\
                               & dangerzone          & False          & \(-0.001\)       & \textbf{687.37 \(\pm\) 16.75} & \textbf{0.63 \(\pm\) 0.01}   & 0.00 \(\pm\) 0.00             & 604.86 \(\pm\) 44.86            & 0.62 \(\pm\) 0.04          & 0.00 \(\pm\) 0.00            \\
                               & dithering           & False          & \(-0.001\)       & \textbf{640.35 \(\pm\) 25.94} & \textbf{0.67 \(\pm\) 0.09}   & \textbf{0.17 \(\pm\) 0.00027} & 604.86 \(\pm\) 44.86            & 0.62 \(\pm\) 0.04          & 0.53 \(\pm\) 0.0064          \\
\bottomrule
\end{tabular}

%% file: mujoco-results.tex
%
\begin{tabular}{cccccccccccccc}
\toprule
\multicolumn{1}{c}{\multirow{3}{*}{Environment} }
& \multicolumn{1}{c}{\multirow{3}{*}{Constraint} }
& \multicolumn{2}{c}{\multirow{2}{*}{Baseline}}
& \multicolumn{10}{c}{Reward Shaping Value} \\
\cmidrule(lr){5-14}
& & \multicolumn{2}{c}{} & \multicolumn{2}{c}{0} & \multicolumn{2}{c}{$-1$} & \multicolumn{2}{c}{$-10$} & \multicolumn{2}{c}{$-100$} & \multicolumn{2}{c}{$-1000$} \\
\cmidrule(lr){3-14}
\multicolumn{1}{c}{} & & rewards & violations & rewards & violations & rewards & violations & rewards & violations & rewards & violations & rewards & violations  \\
\midrule
\multicolumn{1}{c}{\multirow{1}{*}{Half cheetah} } &
\multicolumn{1}{c}{dithering} & \rv{1555.30}{27.42} & \rv{82.84}{6.26} & \rv{1458.68}{32.23} & \rv{80.57}{5.74}  & \rv{2054.84}{451.78} & \rv{73.06}{13.37}  & \hlv{\rv{2524.10}{436.68}} & \rv{62.31}{11.25}  & \rv{1495.21}{165.21} & \rv{43.27}{10.21}  & \rv{639.00}{30.38} & \hlv{\rv{16.73}{6.70}} \\
\multicolumn{1}{c}{\multirow{2}{*}{Reacher} } &
\multicolumn{1}{c}{actuation} & \multicolumn{1}{c}{\multirow{2}{*}{\rv{-6.55}{0.94}}} & \rv{0.61}{0.06} & \rv{-6.28}{0.51} & \rv{0.59}{0.04}  & \rv{-6.55}{0.98} & \rv{0.02}{0.03}  & \hlv{\rv{-5.28}{0.22}} & \hlv{\rv{0.00}{0.00}}  & \rv{-8.36}{0.40} & \hlv{\rv{0.00}{0.00}}  & \rv{-13.44}{0.61} & \hlv{\rv{0.00}{0.00}} \\
\multicolumn{1}{c}{}                        &
                                              \multicolumn{1}{c}{dynamic actuation} & \multicolumn{1}{c}{} & \hlv{\rv{0.00}{0.00}} & \rv{-5.93}{1.67} & \hlv{\rv{0.00}{0.00}}  & \rv{-5.69}{1.02} & \hlv{\rv{0.00}{0.00}}  & \rv{-5.53}{1.32} & \hlv{\rv{0.00}{0.00}}  & \hlv{\rv{-4.75}{0.88}} & \hlv{\rv{0.00}{0.00}}  & \rv{-11.40}{0.61} & \hlv{\rv{0.00}{0.00}} \\
  \bottomrule
\end{tabular}

%% file: hard-results.tex
\begin{tabular}{llcccccc}
\toprule
                               &                     & \multicolumn{2}{c}{\textbf{Training and Evaluation}} & \multicolumn{2}{c}{\textbf{Training Only}} & \multicolumn{2}{c}{\textbf{Evaluation Only}}                                                                  \\
\cmidrule(lr){3-4}\cmidrule(lr){5-6}\cmidrule(lr){7-8}
                               &                     & \textbf{Mean}                                        & \textbf{Mean}                              & \textbf{Mean}                 & \textbf{Mean}            & \textbf{Mean}           & \textbf{Mean}            \\
\textbf{Environment}           & \textbf{Constraint} & \textbf{Episode Reward}                              & \textbf{Viols/100 Steps}                   & \textbf{Episode Reward}       & \textbf{Viols/100 Steps} & \textbf{Episode Reward} & \textbf{Viols/100 Steps} \\
\midrule
\multirow{3}{*}{Breakout}      & actuation           & 302.00 \(\pm\) 20.75                                 & 0.0 \(\pm\) 0.0                            & \textbf{320.31 \(\pm\) 6.09}  & 0.092 \(\pm\) 0.00026    & 314.91 \(\pm\) 13.80    & 0.0 \(\pm\) 0.0          \\
                               & dithering           & 295.31 \(\pm\) 29.07                                 & 0.0 \(\pm\) 0.0                            & \textbf{276.72 \(\pm\) 15.50} & 0.0073 \(\pm\) 3.1e-05   & 275.25 \(\pm\) 12.67    & 0.0 \(\pm\) 0.0          \\
                               & paddle ball         & 218.14 \(\pm\) 22.85                                 & 0.0 \(\pm\) 0.0                            & \textbf{281.77 \(\pm\) 11.03} & 0.11 \(\pm\) 0.00029     & 229.00 \(\pm\) 11.65    & 0.0 \(\pm\) 0.0          \\
\midrule
\multirow{2}{*}{Seaquest}      & actuation           & \textbf{1926.97 \(\pm\) 430.24}                      & 0.0 \(\pm\) 0.0                            & 1899.78 \(\pm\) 502.27        & 8.30 \(\pm\) 0.043       & 1895.77 \(\pm\) 366.79  & 0.0 \(\pm\) 0.0          \\
                               & dithering           & \textbf{2284.78 \(\pm\) 15.45}                       & 0.0 \(\pm\) 0.0                            & 2256.06 \(\pm\) 30.53         & 0.01 \(\pm\) 2.8e-05     & 2267.53 \(\pm\) 23.94   & 0.0 \(\pm\) 0.0          \\
\midrule
\multirow{3}{*}{SpaceInvaders} & actuation           & \textbf{586.66 \(\pm\) 58.69}                        & 0.0 \(\pm\) 0.0                            & 582.79 \(\pm\) 51.47          & 14.62 \(\pm\) 0.012      & 583.13 \(\pm\) 60.60    & 0.0 \(\pm\) 0.0          \\
                               & dangerzone          & 613.61 \(\pm\) 24.05                                 & 0.0 \(\pm\) 0.0                            & \textbf{733.52 \(\pm\) 16.95} & 0.0 \(\pm\) 0.0          & 613.82 \(\pm\) 27.52    & 0.0 \(\pm\) 0.0          \\
                               & dithering           & \textbf{627.40 \(\pm\) 31.43}                        & 0.0 \(\pm\) 0.0                            & 624.33 \(\pm\) 25.27          & 0.008 \(\pm\) 3e-05      & 626.59 \(\pm\) 31.04    & 0.0 \(\pm\) 0.0          \\
\bottomrule
\end{tabular}

%% file: seaquest_2d_regex.tex
The no-2d-dithering constraint used with the Seaquest environment was generated by a simple Python script which generated all sequences of up to 4 moves and filtered them to those which end where they begin with no side effects (e.g., pressing the fire button).

\texttt{\seqsplit{((2|A)(2|A)(5|D)(5|D))|((2|A)(5|D)(2|A)(5|D))|((2|A)(5|D)(5|D)(2|A))|((5|D)(2|A)(2|A)(5|D))|((5|D)(2|A)(5|D)(2|A))|((5|D)(5|D)(2|A)(2|A))|((2|A)(2|A)(8|G)(9|H))|((2|A)(2|A)(9|H)(8|G))|((2|A)(8|G)(2|A)(9|H))|((2|A)(8|G)(9|H)(2|A))|((2|A)(9|H)(2|A)(8|G))|((2|A)(9|H)(8|G)(2|A))|((8|G)(2|A)(2|A)(9|H))|((8|G)(2|A)(9|H)(2|A))|((8|G)(9|H)(2|A)(2|A))|((9|H)(2|A)(2|A)(8|G))|((9|H)(2|A)(8|G)(2|A))|((9|H)(8|G)(2|A)(2|A))|((2|A)(3|B)(4|C)(5|D))|((2|A)(3|B)(5|D)(4|C))|((2|A)(4|C)(3|B)(5|D))|((2|A)(4|C)(5|D)(3|B))|((2|A)(5|D)(3|B)(4|C))|((2|A)(5|D)(4|C)(3|B))|((3|B)(2|A)(4|C)(5|D))|((3|B)(2|A)(5|D)(4|C))|((3|B)(4|C)(2|A)(5|D))|((3|B)(4|C)(5|D)(2|A))|((3|B)(5|D)(2|A)(4|C))|((3|B)(5|D)(4|C)(2|A))|((4|C)(2|A)(3|B)(5|D))|((4|C)(2|A)(5|D)(3|B))|((4|C)(3|B)(2|A)(5|D))|((4|C)(3|B)(5|D)(2|A))|((4|C)(5|D)(2|A)(3|B))|((4|C)(5|D)(3|B)(2|A))|((5|D)(2|A)(3|B)(4|C))|((5|D)(2|A)(4|C)(3|B))|((5|D)(3|B)(2|A)(4|C))|((5|D)(3|B)(4|C)(2|A))|((5|D)(4|C)(2|A)(3|B))|((5|D)(4|C)(3|B)(2|A))|((2|A)(3|B)(9|H))|((2|A)(3|B)(9|H))|((2|A)(9|H)(3|B))|((2|A)(9|H)(3|B))|((2|A)(3|B)(9|H))|((2|A)(9|H)(3|B))|((3|B)(2|A)(9|H))|((3|B)(2|A)(9|H))|((3|B)(9|H)(2|A))|((3|B)(9|H)(2|A))|((3|B)(2|A)(9|H))|((3|B)(9|H)(2|A))|((9|H)(2|A)(3|B))|((9|H)(2|A)(3|B))|((9|H)(3|B)(2|A))|((9|H)(3|B)(2|A))|((9|H)(2|A)(3|B))|((9|H)(3|B)(2|A))|((2|A)(3|B)(9|H))|((2|A)(9|H)(3|B))|((3|B)(2|A)(9|H))|((3|B)(9|H)(2|A))|((9|H)(2|A)(3|B))|((9|H)(3|B)(2|A))|((2|A)(4|C)(8|G))|((2|A)(4|C)(8|G))|((2|A)(8|G)(4|C))|((2|A)(8|G)(4|C))|((2|A)(4|C)(8|G))|((2|A)(8|G)(4|C))|((4|C)(2|A)(8|G))|((4|C)(2|A)(8|G))|((4|C)(8|G)(2|A))|((4|C)(8|G)(2|A))|((4|C)(2|A)(8|G))|((4|C)(8|G)(2|A))|((8|G)(2|A)(4|C))|((8|G)(2|A)(4|C))|((8|G)(4|C)(2|A))|((8|G)(4|C)(2|A))|((8|G)(2|A)(4|C))|((8|G)(4|C)(2|A))|((2|A)(4|C)(8|G))|((2|A)(8|G)(4|C))|((4|C)(2|A)(8|G))|((4|C)(8|G)(2|A))|((8|G)(2|A)(4|C))|((8|G)(4|C)(2|A))|((2|A)(5|D)(6|E)(9|H))|((2|A)(5|D)(9|H)(6|E))|((2|A)(6|E)(5|D)(9|H))|((2|A)(6|E)(9|H)(5|D))|((2|A)(9|H)(5|D)(6|E))|((2|A)(9|H)(6|E)(5|D))|((5|D)(2|A)(6|E)(9|H))|((5|D)(2|A)(9|H)(6|E))|((5|D)(6|E)(2|A)(9|H))|((5|D)(6|E)(9|H)(2|A))|((5|D)(9|H)(2|A)(6|E))|((5|D)(9|H)(6|E)(2|A))|((6|E)(2|A)(5|D)(9|H))|((6|E)(2|A)(9|H)(5|D))|((6|E)(5|D)(2|A)(9|H))|((6|E)(5|D)(9|H)(2|A))|((6|E)(9|H)(2|A)(5|D))|((6|E)(9|H)(5|D)(2|A))|((9|H)(2|A)(5|D)(6|E))|((9|H)(2|A)(6|E)(5|D))|((9|H)(5|D)(2|A)(6|E))|((9|H)(5|D)(6|E)(2|A))|((9|H)(6|E)(2|A)(5|D))|((9|H)(6|E)(5|D)(2|A))|((2|A)(5|D)(7|F)(8|G))|((2|A)(5|D)(8|G)(7|F))|((2|A)(7|F)(5|D)(8|G))|((2|A)(7|F)(8|G)(5|D))|((2|A)(8|G)(5|D)(7|F))|((2|A)(8|G)(7|F)(5|D))|((5|D)(2|A)(7|F)(8|G))|((5|D)(2|A)(8|G)(7|F))|((5|D)(7|F)(2|A)(8|G))|((5|D)(7|F)(8|G)(2|A))|((5|D)(8|G)(2|A)(7|F))|((5|D)(8|G)(7|F)(2|A))|((7|F)(2|A)(5|D)(8|G))|((7|F)(2|A)(8|G)(5|D))|((7|F)(5|D)(2|A)(8|G))|((7|F)(5|D)(8|G)(2|A))|((7|F)(8|G)(2|A)(5|D))|((7|F)(8|G)(5|D)(2|A))|((8|G)(2|A)(5|D)(7|F))|((8|G)(2|A)(7|F)(5|D))|((8|G)(5|D)(2|A)(7|F))|((8|G)(5|D)(7|F)(2|A))|((8|G)(7|F)(2|A)(5|D))|((8|G)(7|F)(5|D)(2|A))|((2|A)(5|D))|((2|A)(5|D))|((2|A)(5|D))|((5|D)(2|A))|((5|D)(2|A))|((5|D)(2|A))|((2|A)(5|D))|((2|A)(5|D))|((5|D)(2|A))|((5|D)(2|A))|((2|A)(5|D))|((5|D)(2|A))|((3|B)(3|B)(4|C)(4|C))|((3|B)(4|C)(3|B)(4|C))|((3|B)(4|C)(4|C)(3|B))|((4|C)(3|B)(3|B)(4|C))|((4|C)(3|B)(4|C)(3|B))|((4|C)(4|C)(3|B)(3|B))|((3|B)(3|B)(7|F)(9|H))|((3|B)(3|B)(9|H)(7|F))|((3|B)(7|F)(3|B)(9|H))|((3|B)(7|F)(9|H)(3|B))|((3|B)(9|H)(3|B)(7|F))|((3|B)(9|H)(7|F)(3|B))|((7|F)(3|B)(3|B)(9|H))|((7|F)(3|B)(9|H)(3|B))|((7|F)(9|H)(3|B)(3|B))|((9|H)(3|B)(3|B)(7|F))|((9|H)(3|B)(7|F)(3|B))|((9|H)(7|F)(3|B)(3|B))|((3|B)(4|C)(6|E)(9|H))|((3|B)(4|C)(9|H)(6|E))|((3|B)(6|E)(4|C)(9|H))|((3|B)(6|E)(9|H)(4|C))|((3|B)(9|H)(4|C)(6|E))|((3|B)(9|H)(6|E)(4|C))|((4|C)(3|B)(6|E)(9|H))|((4|C)(3|B)(9|H)(6|E))|((4|C)(6|E)(3|B)(9|H))|((4|C)(6|E)(9|H)(3|B))|((4|C)(9|H)(3|B)(6|E))|((4|C)(9|H)(6|E)(3|B))|((6|E)(3|B)(4|C)(9|H))|((6|E)(3|B)(9|H)(4|C))|((6|E)(4|C)(3|B)(9|H))|((6|E)(4|C)(9|H)(3|B))|((6|E)(9|H)(3|B)(4|C))|((6|E)(9|H)(4|C)(3|B))|((9|H)(3|B)(4|C)(6|E))|((9|H)(3|B)(6|E)(4|C))|((9|H)(4|C)(3|B)(6|E))|((9|H)(4|C)(6|E)(3|B))|((9|H)(6|E)(3|B)(4|C))|((9|H)(6|E)(4|C)(3|B))|((3|B)(4|C)(7|F)(8|G))|((3|B)(4|C)(8|G)(7|F))|((3|B)(7|F)(4|C)(8|G))|((3|B)(7|F)(8|G)(4|C))|((3|B)(8|G)(4|C)(7|F))|((3|B)(8|G)(7|F)(4|C))|((4|C)(3|B)(7|F)(8|G))|((4|C)(3|B)(8|G)(7|F))|((4|C)(7|F)(3|B)(8|G))|((4|C)(7|F)(8|G)(3|B))|((4|C)(8|G)(3|B)(7|F))|((4|C)(8|G)(7|F)(3|B))|((7|F)(3|B)(4|C)(8|G))|((7|F)(3|B)(8|G)(4|C))|((7|F)(4|C)(3|B)(8|G))|((7|F)(4|C)(8|G)(3|B))|((7|F)(8|G)(3|B)(4|C))|((7|F)(8|G)(4|C)(3|B))|((8|G)(3|B)(4|C)(7|F))|((8|G)(3|B)(7|F)(4|C))|((8|G)(4|C)(3|B)(7|F))|((8|G)(4|C)(7|F)(3|B))|((8|G)(7|F)(3|B)(4|C))|((8|G)(7|F)(4|C)(3|B))|((3|B)(4|C))|((3|B)(4|C))|((3|B)(4|C))|((4|C)(3|B))|((4|C)(3|B))|((4|C)(3|B))|((3|B)(4|C))|((3|B)(4|C))|((4|C)(3|B))|((4|C)(3|B))|((3|B)(4|C))|((4|C)(3|B))|((3|B)(5|D)(7|F))|((3|B)(5|D)(7|F))|((3|B)(7|F)(5|D))|((3|B)(7|F)(5|D))|((3|B)(5|D)(7|F))|((3|B)(7|F)(5|D))|((5|D)(3|B)(7|F))|((5|D)(3|B)(7|F))|((5|D)(7|F)(3|B))|((5|D)(7|F)(3|B))|((5|D)(3|B)(7|F))|((5|D)(7|F)(3|B))|((7|F)(3|B)(5|D))|((7|F)(3|B)(5|D))|((7|F)(5|D)(3|B))|((7|F)(5|D)(3|B))|((7|F)(3|B)(5|D))|((7|F)(5|D)(3|B))|((3|B)(5|D)(7|F))|((3|B)(7|F)(5|D))|((5|D)(3|B)(7|F))|((5|D)(7|F)(3|B))|((7|F)(3|B)(5|D))|((7|F)(5|D)(3|B))|((4|C)(4|C)(6|E)(8|G))|((4|C)(4|C)(8|G)(6|E))|((4|C)(6|E)(4|C)(8|G))|((4|C)(6|E)(8|G)(4|C))|((4|C)(8|G)(4|C)(6|E))|((4|C)(8|G)(6|E)(4|C))|((6|E)(4|C)(4|C)(8|G))|((6|E)(4|C)(8|G)(4|C))|((6|E)(8|G)(4|C)(4|C))|((8|G)(4|C)(4|C)(6|E))|((8|G)(4|C)(6|E)(4|C))|((8|G)(6|E)(4|C)(4|C))|((4|C)(5|D)(6|E))|((4|C)(5|D)(6|E))|((4|C)(6|E)(5|D))|((4|C)(6|E)(5|D))|((4|C)(5|D)(6|E))|((4|C)(6|E)(5|D))|((5|D)(4|C)(6|E))|((5|D)(4|C)(6|E))|((5|D)(6|E)(4|C))|((5|D)(6|E)(4|C))|((5|D)(4|C)(6|E))|((5|D)(6|E)(4|C))|((6|E)(4|C)(5|D))|((6|E)(4|C)(5|D))|((6|E)(5|D)(4|C))|((6|E)(5|D)(4|C))|((6|E)(4|C)(5|D))|((6|E)(5|D)(4|C))|((4|C)(5|D)(6|E))|((4|C)(6|E)(5|D))|((5|D)(4|C)(6|E))|((5|D)(6|E)(4|C))|((6|E)(4|C)(5|D))|((6|E)(5|D)(4|C))|((5|D)(5|D)(6|E)(7|F))|((5|D)(5|D)(7|F)(6|E))|((5|D)(6|E)(5|D)(7|F))|((5|D)(6|E)(7|F)(5|D))|((5|D)(7|F)(5|D)(6|E))|((5|D)(7|F)(6|E)(5|D))|((6|E)(5|D)(5|D)(7|F))|((6|E)(5|D)(7|F)(5|D))|((6|E)(7|F)(5|D)(5|D))|((7|F)(5|D)(5|D)(6|E))|((7|F)(5|D)(6|E)(5|D))|((7|F)(6|E)(5|D)(5|D))|((6|E)(6|E)(9|H)(9|H))|((6|E)(9|H)(6|E)(9|H))|((6|E)(9|H)(9|H)(6|E))|((9|H)(6|E)(6|E)(9|H))|((9|H)(6|E)(9|H)(6|E))|((9|H)(9|H)(6|E)(6|E))|((6|E)(7|F)(8|G)(9|H))|((6|E)(7|F)(9|H)(8|G))|((6|E)(8|G)(7|F)(9|H))|((6|E)(8|G)(9|H)(7|F))|((6|E)(9|H)(7|F)(8|G))|((6|E)(9|H)(8|G)(7|F))|((7|F)(6|E)(8|G)(9|H))|((7|F)(6|E)(9|H)(8|G))|((7|F)(8|G)(6|E)(9|H))|((7|F)(8|G)(9|H)(6|E))|((7|F)(9|H)(6|E)(8|G))|((7|F)(9|H)(8|G)(6|E))|((8|G)(6|E)(7|F)(9|H))|((8|G)(6|E)(9|H)(7|F))|((8|G)(7|F)(6|E)(9|H))|((8|G)(7|F)(9|H)(6|E))|((8|G)(9|H)(6|E)(7|F))|((8|G)(9|H)(7|F)(6|E))|((9|H)(6|E)(7|F)(8|G))|((9|H)(6|E)(8|G)(7|F))|((9|H)(7|F)(6|E)(8|G))|((9|H)(7|F)(8|G)(6|E))|((9|H)(8|G)(6|E)(7|F))|((9|H)(8|G)(7|F)(6|E))|((6|E)(9|H))|((6|E)(9|H))|((6|E)(9|H))|((9|H)(6|E))|((9|H)(6|E))|((9|H)(6|E))|((6|E)(9|H))|((6|E)(9|H))|((9|H)(6|E))|((9|H)(6|E))|((6|E)(9|H))|((9|H)(6|E))|((7|F)(7|F)(8|G)(8|G))|((7|F)(8|G)(7|F)(8|G))|((7|F)(8|G)(8|G)(7|F))|((8|G)(7|F)(7|F)(8|G))|((8|G)(7|F)(8|G)(7|F))|((8|G)(8|G)(7|F)(7|F))|((7|F)(8|G))|((7|F)(8|G))|((7|F)(8|G))|((8|G)(7|F))|((8|G)(7|F))|((8|G)(7|F))|((7|F)(8|G))|((7|F)(8|G))|((8|G)(7|F))|((8|G)(7|F))|((7|F)(8|G))|((8|G)(7|F))}}

%% file: main.bbl
\begin{thebibliography}{38}
\providecommand{\natexlab}[1]{#1}
\providecommand{\url}[1]{\texttt{#1}}
\expandafter\ifx\csname urlstyle\endcsname\relax
  \providecommand{\doi}[1]{doi: #1}\else
  \providecommand{\doi}{doi: \begingroup \urlstyle{rm}\Url}\fi

\bibitem[Alshiekh et~al.(2018)Alshiekh, Bloem, Ehlers, K{\"o}nighofer, Niekum,
  and Topcu]{2018shieldedAAAI}
Mohammed Alshiekh, Roderick Bloem, R{\"u}diger Ehlers, Bettina K{\"o}nighofer,
  Scott Niekum, and Ufuk Topcu.
\newblock Safe reinforcement learning via shielding.
\newblock In \emph{Proceedings of AAAI 18}, pp.\  2669--2678, 2018.

\bibitem[Altman(1999)]{altman1999constrained}
Eitan Altman.
\newblock \emph{Constrained Markov decision processes}, volume~7.
\newblock CRC Press, 1999.

\bibitem[Amodei et~al.(2016)Amodei, Olah, Steinhardt, Christiano, Schulman, and
  Man{\'e}]{amodei2016concrete}
Dario Amodei, Chris Olah, Jacob Steinhardt, Paul Christiano, John Schulman, and
  Dan Man{\'e}.
\newblock Concrete problems in {AI} safety.
\newblock \emph{arXiv preprint arXiv:1606.06565}, 2016.

\bibitem[Aréchiga \& Krogh(2014)Aréchiga and Krogh]{arechiga2014using}
N.~Aréchiga and B.~Krogh.
\newblock Using verified control envelopes for safe controller design.
\newblock In \emph{2014 American Control Conference}, pp.\  2918--2923, June
  2014.

\bibitem[Baier et~al.(2003)Baier, Haverkort, Hermanns, and
  Katoen]{baier2003model}
Christel Baier, Boudewijn Haverkort, Holger Hermanns, and Joost-Pieter Katoen.
\newblock Model-checking algorithms for continuous-time {Markov} chains.
\newblock \emph{IEEE Transactions on software engineering}, \penalty0
  (6):\penalty0 524--541, 2003.

\bibitem[{Bellemare} et~al.(2013){Bellemare}, {Naddaf}, {Veness}, and
  {Bowling}]{bellemare13arcade}
M.~G. {Bellemare}, Y.~{Naddaf}, J.~{Veness}, and M.~{Bowling}.
\newblock The arcade learning environment: An evaluation platform for general
  agents.
\newblock \emph{Journal of Artificial Intelligence Research}, 47:\penalty0
  253--279, jun 2013.

\bibitem[Bouajjani et~al.(1997)Bouajjani, Esparza, and
  Maler]{bouajjani1997reachability}
Ahmed Bouajjani, Javier Esparza, and Oded Maler.
\newblock Reachability analysis of pushdown automata: Application to
  model-checking.
\newblock In \emph{International Conference on Concurrency Theory}, pp.\
  135--150. Springer, 1997.

\bibitem[Brockman et~al.(2016)Brockman, Cheung, Pettersson, Schneider,
  Schulman, Tang, and Zaremba]{brockman2016openai}
Greg Brockman, Vicki Cheung, Ludwig Pettersson, Jonas Schneider, John Schulman,
  Jie Tang, and Wojciech Zaremba.
\newblock {OpenAI} gym.
\newblock \emph{arXiv preprint arXiv:1606.01540}, 2016.

\bibitem[Camacho et~al.(2017{\natexlab{a}})Camacho, Chen, Sanner, and
  McIlraith]{DBLP:conf/socs/CamachoCSM17}
Alberto Camacho, Oscar Chen, Scott Sanner, and Sheila~A. McIlraith.
\newblock Non-markovian rewards expressed in {LTL:} guiding search via reward
  shaping.
\newblock In \emph{Proceedings of the Tenth International Symposium on
  Combinatorial Search, {SOCS} 2017, 16-17 June 2017, Pittsburgh, Pennsylvania,
  {USA.}}, pp.\  159--160, 2017{\natexlab{a}}.

\bibitem[Camacho et~al.(2017{\natexlab{b}})Camacho, Chen, Sanner, and
  McIlraith]{camacho2017decision}
Alberto Camacho, Oscar Chen, Scott Sanner, and Sheila~A McIlraith.
\newblock Decision-making with non-markovian rewards: From {LTL} to
  automata-based reward shaping.
\newblock In \emph{Proceedings of the Multi-disciplinary Conference on
  Reinforcement Learning and Decision Making (RLDM)}, pp.\  279--283,
  2017{\natexlab{b}}.

\bibitem[Chen \& Ro\c{s}u(2007)Chen and Ro\c{s}u]{Chen:2007:MOP}
Feng Chen and Grigore Ro\c{s}u.
\newblock Mop: An efficient and generic runtime verification framework.
\newblock In \emph{Proceedings of the 22Nd Annual ACM SIGPLAN Conference on
  Object-oriented Programming Systems and Applications}, OOPSLA '07, 2007.

\bibitem[Chow et~al.(2017)Chow, Ghavamzadeh, Janson, and Pavone]{chow2017risk}
Yinlam Chow, Mohammad Ghavamzadeh, Lucas Janson, and Marco Pavone.
\newblock Risk-constrained reinforcement learning with percentile risk
  criteria.
\newblock \emph{The Journal of Machine Learning Research}, 18\penalty0
  (1):\penalty0 6070--6120, 2017.

\bibitem[Clarke et~al.(2001)Clarke, Grumberg, and
  Peled]{DBLP:books/daglib/0007403}
Edmund~M. Clarke, Orna Grumberg, and Doron~A. Peled.
\newblock \emph{Model checking}.
\newblock {MIT} Press, 2001.
\newblock ISBN 978-0-262-03270-4.

\bibitem[Dalal et~al.(2018)Dalal, Dvijotham, Vecerik, Hester, Paduraru, and
  Tassa]{dalal2018safe}
Gal Dalal, Krishnamurthy Dvijotham, Matej Vecerik, Todd Hester, Cosmin
  Paduraru, and Yuval Tassa.
\newblock Safe exploration in continuous action spaces.
\newblock \emph{arXiv preprint arXiv:1801.08757}, 2018.

\bibitem[De~Giacomo et~al.(2020)De~Giacomo, Favorito, Iocchi, Patrizi, and
  Ronca]{detemporal}
Giuseppe De~Giacomo, Marco Favorito, Luca Iocchi, Fabio Patrizi, and Alessandro
  Ronca.
\newblock Temporal logic monitoring rewards via transducers.
\newblock 2020.

\bibitem[Dhariwal et~al.(2017)Dhariwal, Hesse, Klimov, Nichol, Plappert,
  Radford, Schulman, Sidor, Wu, and Zhokhov]{baselines}
Prafulla Dhariwal, Christopher Hesse, Oleg Klimov, Alex Nichol, Matthias
  Plappert, Alec Radford, John Schulman, Szymon Sidor, Yuhuai Wu, and Peter
  Zhokhov.
\newblock {OpenAI} baselines.
\newblock \url{https://github.com/openai/baselines}, 2017.

\bibitem[Hasanbeig et~al.(2018)Hasanbeig, Abate, and
  Kroening]{DBLP:journals/corr/abs-1809-07823}
Mohammadhosein Hasanbeig, Alessandro Abate, and Daniel Kroening.
\newblock Logically-constrained neural fitted q-iteration.
\newblock \emph{CoRR}, abs/1809.07823, 2018.

\bibitem[Helm et~al.(1990)Helm, Holland, and Gangopadhyay]{helm1990contracts}
Richard Helm, Ian~M Holland, and Dipayan Gangopadhyay.
\newblock Contracts: specifying behavioral compositions in object-oriented
  systems.
\newblock \emph{ACM Sigplan Notices}, 25\penalty0 (10):\penalty0 169--180,
  1990.

\bibitem[Huth \& Kwiatkowska(1997)Huth and Kwiatkowska]{DBLP:conf/lics/HuthK97}
Michael Huth and Marta~Z. Kwiatkowska.
\newblock Quantitative analysis and model checking.
\newblock In \emph{Proceedings, 12th Annual {IEEE} Symposium on Logic in
  Computer Science, Warsaw, Poland, June 29 - July 2, 1997}, pp.\  111--122,
  1997.

\bibitem[Icarte et~al.(2018{\natexlab{a}})Icarte, Klassen, Valenzano, and
  McIlraith]{DBLP:conf/atal/IcarteKVM18}
Rodrigo~Toro Icarte, Toryn~Q. Klassen, Richard~Anthony Valenzano, and Sheila~A.
  McIlraith.
\newblock Teaching multiple tasks to an {RL} agent using {LTL}.
\newblock In \emph{Proceedings of the 17th International Conference on
  Autonomous Agents and MultiAgent Systems, {AAMAS} 2018, Stockholm, Sweden,
  July 10-15, 2018}, pp.\  452--461, 2018{\natexlab{a}}.

\bibitem[Icarte et~al.(2018{\natexlab{b}})Icarte, Klassen, Valenzano, and
  McIlraith]{DBLP:conf/icml/IcarteKVM18}
Rodrigo~Toro Icarte, Toryn~Q. Klassen, Richard~Anthony Valenzano, and Sheila~A.
  McIlraith.
\newblock Using reward machines for high-level task specification and
  decomposition in reinforcement learning.
\newblock In \emph{Proceedings of the 35th International Conference on Machine
  Learning, {ICML} 2018, Stockholmsm{\"{a}}ssan, Stockholm, Sweden, July 10-15,
  2018}, pp.\  2112--2121, 2018{\natexlab{b}}.

\bibitem[Jansen et~al.(2018)Jansen, K{\"o}nighofer, Junges, and
  Bloem]{jansen2018shielded}
Nils Jansen, Bettina K{\"o}nighofer, Sebastian Junges, and Roderick Bloem.
\newblock Shielded decision-making in mdps.
\newblock \emph{arXiv preprint arXiv:1807.06096}, 2018.

\bibitem[Kupferman et~al.(2000)Kupferman, Vardi, and
  Wolper]{kupferman2000automata}
Orna Kupferman, Moshe~Y Vardi, and Pierre Wolper.
\newblock An automata-theoretic approach to branching-time model checking.
\newblock \emph{Journal of the ACM (JACM)}, 47\penalty0 (2):\penalty0 312--360,
  2000.

\bibitem[Kwiatkowska et~al.(2002)Kwiatkowska, Norman, and
  Parker]{kwiatkowska2002prism}
Marta Kwiatkowska, Gethin Norman, and David Parker.
\newblock Prism: Probabilistic symbolic model checker.
\newblock In \emph{International Conference on Modelling Techniques and Tools
  for Computer Performance Evaluation}, pp.\  200--204. Springer, 2002.

\bibitem[Li et~al.(2017)Li, Vasile, and Belta]{DBLP:conf/iros/LiVB17}
Xiao Li, Cristian~Ioan Vasile, and Calin Belta.
\newblock Reinforcement learning with temporal logic rewards.
\newblock In \emph{2017 {IEEE/RSJ} International Conference on Intelligent
  Robots and Systems, {IROS} 2017, Vancouver, BC, Canada, September 24-28,
  2017}, pp.\  3834--3839, 2017.

\bibitem[Li et~al.(2018)Li, Ma, and Belta]{DBLP:journals/corr/abs-1809-06305}
Xiao Li, Yao Ma, and Calin Belta.
\newblock Automata guided reinforcement learning with demonstrations.
\newblock \emph{CoRR}, abs/1809.06305, 2018.

\bibitem[Littman et~al.(2017)Littman, Topcu, Fu, Jr., Wen, and
  MacGlashan]{DBLP:journals/corr/LittmanTFIWM17}
Michael~L. Littman, Ufuk Topcu, Jie Fu, Charles Lee~Isbell Jr., Min Wen, and
  James MacGlashan.
\newblock Environment-independent task specifications via {GLTL}.
\newblock \emph{CoRR}, abs/1704.04341, 2017.

\bibitem[Meyer(1992)]{meyer1992applying}
Bertrand Meyer.
\newblock Applying `design by contract'.
\newblock \emph{Computer}, 25\penalty0 (10):\penalty0 40--51, 1992.

\bibitem[Mousavi et~al.(2014)Mousavi, Ghazanfari, Mozayani, and
  Jahed{-}Motlagh]{DBLP:journals/asc/MousaviGMJ14}
Seyed~Sajad Mousavi, Behzad Ghazanfari, Nasser Mozayani, and Mohammad~Reza
  Jahed{-}Motlagh.
\newblock Automatic abstraction controller in reinforcement learning agent via
  automata.
\newblock \emph{Appl. Soft Comput.}, 25:\penalty0 118--128, 2014.

\bibitem[Ng et~al.(1999)Ng, Harada, and Russell]{DBLP:conf/icml/NgHR99}
Andrew~Y. Ng, Daishi Harada, and Stuart~J. Russell.
\newblock Policy invariance under reward transformations: Theory and
  application to reward shaping.
\newblock In \emph{Proceedings of the Sixteenth International Conference on
  Machine Learning {(ICML} 1999), Bled, Slovenia, June 27 - 30, 1999}, pp.\
  278--287, 1999.

\bibitem[Pham et~al.(2018)Pham, De~Magistris, and Tachibana]{pham2018optlayer}
Tu-Hoa Pham, Giovanni De~Magistris, and Ryuki Tachibana.
\newblock Optlayer-practical constrained optimization for deep reinforcement
  learning in the real world.
\newblock In \emph{2018 IEEE International Conference on Robotics and
  Automation (ICRA)}, pp.\  6236--6243. IEEE, 2018.

\bibitem[Ray et~al.(2019)Ray, Achiam, and Amodei]{ray2019benchmarking}
Alex Ray, Joshua Achiam, and Dario Amodei.
\newblock Benchmarking safe exploration in deep reinforcement learning.
\newblock \emph{arXiv preprint arXiv:1910.01708}, 2019.

\bibitem[Saunders et~al.(2017)Saunders, Sastry, Stuhlmueller, and
  Evans]{saunders2017trial}
William Saunders, Girish Sastry, Andreas Stuhlmueller, and Owain Evans.
\newblock Trial without error: Towards safe reinforcement learning via human
  intervention.
\newblock \emph{arXiv preprint arXiv:1707.05173}, 2017.

\bibitem[Schulman et~al.(2017)Schulman, Wolski, Dhariwal, Radford, and
  Klimov]{schulman2017proximal}
John Schulman, Filip Wolski, Prafulla Dhariwal, Alec Radford, and Oleg Klimov.
\newblock Proximal policy optimization algorithms.
\newblock \emph{arXiv preprint arXiv:1707.06347}, 2017.

\bibitem[Sutton \& Barto(2018)Sutton and Barto]{sutton2018reinforcement}
Richard~S Sutton and Andrew~G Barto.
\newblock \emph{Reinforcement learning: An introduction}.
\newblock MIT press, 2018.

\bibitem[van~der Walt et~al.(2014)van~der Walt, {S}ch\"onberger,
  {Nunez-Iglesias}, {B}oulogne, {W}arner, {Y}ager, {G}ouillart, {Y}u, and the
  scikit-image contributors]{scikit-image}
{S}t\'efan van~der Walt, {J}ohannes~{L}. {S}ch\"onberger, {J}uan
  {Nunez-Iglesias}, {F}ran\c{c}ois {B}oulogne, {J}oshua~{D}. {W}arner, {N}eil
  {Y}ager, {E}mmanuelle {G}ouillart, {T}ony {Y}u, and the scikit-image
  contributors.
\newblock scikit-image: image processing in {P}ython.
\newblock \emph{PeerJ}, 2:\penalty0 e453, 6 2014.
\newblock ISSN 2167-8359.
\newblock \doi{10.7717/peerj.453}.
\newblock URL \url{https://doi.org/10.7717/peerj.453}.

\bibitem[Wen et~al.(2017)Wen, Papusha, and Topcu]{DBLP:conf/ijcai/WenPT17}
Min Wen, Ivan Papusha, and Ufuk Topcu.
\newblock Learning from demonstrations with high-level side information.
\newblock In \emph{Proceedings of the Twenty-Sixth International Joint
  Conference on Artificial Intelligence, {IJCAI} 2017, Melbourne, Australia,
  August 19-25, 2017}, pp.\  3055--3061, 2017.

\bibitem[Wiewiora et~al.(2003)Wiewiora, Cottrell, and
  Elkan]{wiewiora2003principled}
Eric Wiewiora, Garrison~W Cottrell, and Charles Elkan.
\newblock Principled methods for advising reinforcement learning agents.
\newblock In \emph{Proceedings of the 20th International Conference on Machine
  Learning (ICML-03)}, pp.\  792--799, 2003.

\end{thebibliography}
